\documentclass[runningheads]{llncs}

\usepackage{accv}

\usepackage{accvabbrv}

\usepackage{graphicx}
\usepackage{booktabs}

\usepackage[accsupp]{axessibility}  

\usepackage{hyperref}

\usepackage{orcidlink}

\begin{document}

\title{Polyp-SES: Automatic Polyp Segmentation with Self-Enriched Semantic Model} 

\titlerunning{Polyp-SES}

\author{Quang Vinh Nguyen\inst{1}\orcidlink{0009-0002-3838-4428} \and
Thanh Hoang Son Vo\inst{1}\orcidlink{0009-0001-3278-727X} \and
Sae-Ryung Kang\inst{2}\orcidlink{0000-0003-0172-5508} \and
Soo-Hyung Kim\inst{1}\orcidlink{0000-0003-3575-5035}}

\authorrunning{Q.V. Nguyen and T.H.S. Vo et al.}

\institute{Chonnam National University, Gwangju, South Korea \\
\email{\{vinhbn28, hoangsonvothanh, shkim\}@jnu.ac.kr}\\ \and
Chonnam National University Hwasun Hospital and Medical School, Hwasun, South Korea \\
\email{campanella9@naver.com}}

\maketitle

\begin{abstract}
Automatic polyp segmentation is crucial for effective diagnosis and treatment in colonoscopy images. Traditional methods encounter significant challenges in accurately delineating polyps due to limitations in feature representation and the handling of variability in polyp appearance. Deep learning techniques, including CNN and Transformer-based methods, have been explored to improve polyp segmentation accuracy. However, existing approaches often neglect additional semantics, restricting their ability to acquire adequate contexts of polyps in colonoscopy images. In this paper, we propose an innovative method named ``Automatic Polyp Segmentation with Self-Enriched Semantic Model'' to address these limitations. First, we extract a sequence of features from an input image and decode high-level features to generate an initial segmentation mask. Using the proposed self-enriched semantic module, we query potential semantics and augment deep features with additional semantics, thereby aiding the model in understanding context more effectively. Extensive experiments show superior segmentation performance of the proposed method against state-of-the-art polyp segmentation baselines across five polyp benchmarks in both superior learning and generalization capabilities. 
\end{abstract}

\section{Introduction}
\label{sec:intro}
Medical image segmentation is the process of delineating regions of interest (ROI) within medical images, such as X-rays, MRI scans, CT scans, or histological slides, into meaningful and distinct anatomical structures. This process can assist clinicians in quantitative analysis, diagnosis, treatment planning, and monitoring of diseases. Traditional image processing techniques\cite{M1,M2,M3,M4} have been widely adopted in the medical image segmentation tasks. These methods primarily rely on handcrafted features potentially limiting their generalizability to complex image structures, noise, or variability in image quality. With the rapid development of deep learning, efficient and reliable segmentation solutions~\cite{TransUnet,M6,M7,M8} was introduced in the domain of the medical image segmentation. 

Polyp segmentation is a critical task in medical imaging aimed at accurately identifying and delineating polyps within endoscopic or colonoscopic images. The challenges of identifying and segmenting polyps in medical images can be summarized for several following primary factors. \textbf{Firstly}, the variability in color, shape, size and texture among polyps poses a significant difficulty to precise segmentation. \textbf{Secondly}, the presence of noise, artifacts, or overlapping structures in the image can obscure polyps, increasing the complexity of the segmentation task. \textbf{Thirdly}, polyps may manifest in diverse positions within the gastrointestinal tract, each with its distinct characteristics, thus contributing to the overall variability and difficulty of the segmentation. 
\begin{center}
    \begin{figure*}[t]
    \begin{center}
     \includegraphics[width=0.8\linewidth]{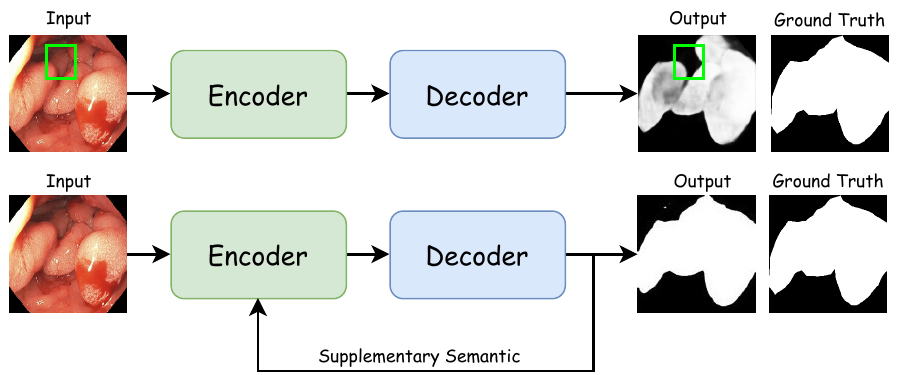}
    \end{center}
    \caption{Deep learning-based automatic polyp segmentation methods often include encoder and decoder parts. Contemporary models struggle to identify and categorize challenging features highlighted within green-bordered areas. This region appears relatively blurry and distinct from the surrounding polyp objects leading to confusion between normal tissues and actual polyps, thereby causing segmentation failures. Providing supplementary semantics promotes the model to obtain comprehensive contextual information about polyp objects, leading to a greatly segmentation performance.}
    \label{motivation}
    \end{figure*}
\end{center} 

Traditional methods~\cite{T1,T2,T3,T4} for polyp segmentation have often faced challenges such as sensitivity to image variations, the need for manual parameter adjustment, limited adaptability, and difficulties in handling noise and artifacts. Efforts leveraging deep learning techniques have been proposed to enhance the effectiveness of automatic polyp segmentation. Firstly, CNN-based methods~\cite{SFA, MSNet, ADSNet, PraNet, U-Net, SANet, gca, fcmd} exploit the ability of neural networks to automatically learn discriminative features and generate precise segmentation results. Despite the considerable success of CNN-based approaches, the limited receptive field poses challenges in capturing global representations. Transformer-based techniques~\cite{SSFormer,Transfuse, Transnetr, MSRAFormer,Polyp-PVT}, on the other hand, excel in capturing global dependencies and long-range contextual information more effectively compared to CNNs, resulting in superior performance in the polyp segmentation task. Nonetheless, transformer-based methodologies encounter difficulties in capturing fine-grained details, which are crucial for accurately detecting and locating polyp objects. In addressing visual comprehension concerns involving semantic segmentation and object recognition, contextual information provides valuable insights that aid in disambiguating similar-looking objects, resolving occlusions, and improving the overall understanding of the visual scene. Notably, the approaches mentioned above primarily relying on poor contextual information neglecting to provide additional semantics. This limitation presents several challenges in comprehending adequate contexts of polyps, which are frequently characterized by noises, ambiguous boundaries, and intricate foregrounds. We have discovered that providing supplementary semantics can assist the model in obtaining comprehensive contextual information about polyp objects, potentially leading to a significant enhancement in segmentation performance as shown in Fig. \ref{motivation}.

Motivated by discussed concerns, our study introduces a new novel approach for automatic polyp segmentation task namely ``Automatic Polyp Segmentation with Self-Enriched Semantic Model''. Initially, we employ an encoder to extract a sequence of multi-scale features. Subsequently, we introduce a Local-to-Global Spatial Fusion (LGSF) mechanism to capture both local and global spatial features before decoding them to generate an initial global feature map. Leveraging the proposed Self-Enriched Semantic (SES) module, we augment deep features with additional semantics, thereby aiding the model in understanding context more effectively. 
Our proposed solution achieves competitive segmentation performance compared to state-of-the-art baselines, showcasing proficiency in learning and generalization capabilities. Notably, it effectively addresses the limitations of prior models when operating in challenging contexts.

\section{Related Work}
\label{sec:related}
\subsection{Automatic Polyp Segmentation.}
Traditional methods~\cite{T1,T2,T3,T4} primarily rely on low-level features such as geometric characteristics, which often result in missed or inaccurate detections due to similarities with neighboring tissues. Recent advancements in deep learning have revolutionized the polyp segmentation by autonomously learning complex features. Among these innovations, U-Net~\cite{U-Net} obtaines significant improvements across various medical imaging tasks By the simplicity and effectiveness design. ACSNet~\cite{ACSNet} refines the conventional skip connections within the U-Net~\cite{U-Net} and selects adaptive features based on a channel attention mechanism. Pranet~\cite{PraNet} utilizes reverse attention mechanisms to refine boundary details in the global feature map through iterative stages enhancing segmentation predictions. MSNet~\cite{MSNet} introduces multi-scale subtraction architecture in order to capture intricate details, eliminate redundancy and complementary information between the multi-scale features. SSformer~\cite{SSFormer} adopts a systematic feature fusion approach, gradually integrating both local and global contextual information, resulting in precise object delineation and boundary detection, while also capturing fine-grained details and comprehensive scene context. Polyp-PVT~\cite{Polyp-PVT} presents a similarity aggregation module to extract local pixel and global semantic cues from the polyp area. 

\subsection{Vision Transformer.}
Transformer~\cite{transformer}, initially successful in NLP, has garnered prominence for their potential in computer vision tasks. Leveraging the transformer mechanism, ViTs~\cite{ViT} effectively captures global dependencies and long-range spatial relationships, enabling comprehensive predictions based on the entire image context. By employing shifted windows instead of fixed-size patches, Swin~\cite{Swin} captures spatial relationships between adjacent patches, leading to enhanced feature representation and learning capabilities. Pyramid Vision Transformer (PVT)~\cite{PVT} incorporates a pyramid feature extraction mechanism to capture multi-scale information from input images. Through the combination of features from various scales, PVT~\cite{PVT} demonstrates proficiency in capturing both local details and global context, facilitating accurate dense prediction. UniFormer~\cite{Uniformer} integrates the strengths of convolutional neural networks (CNNs) and vision transformers (ViTs) into a concise transformer format. This innovative design empowers UniFormer~\cite{Uniformer} to efficiently capture both local redundancies and complex global dependencies, facilitating effective representation learning. MetaFormer~\cite{Metaformer} has recently demonstrated the commendable performance in computer vision tasks. This study meticulously examines various token mixers, spanning from basic operators like identity mapping or global random mixing to established techniques such as separable convolution and vanilla self-attention.

\section{Method}
\begin{center}
    \begin{figure}[t]
    \begin{center}
     \includegraphics[width=1\linewidth]{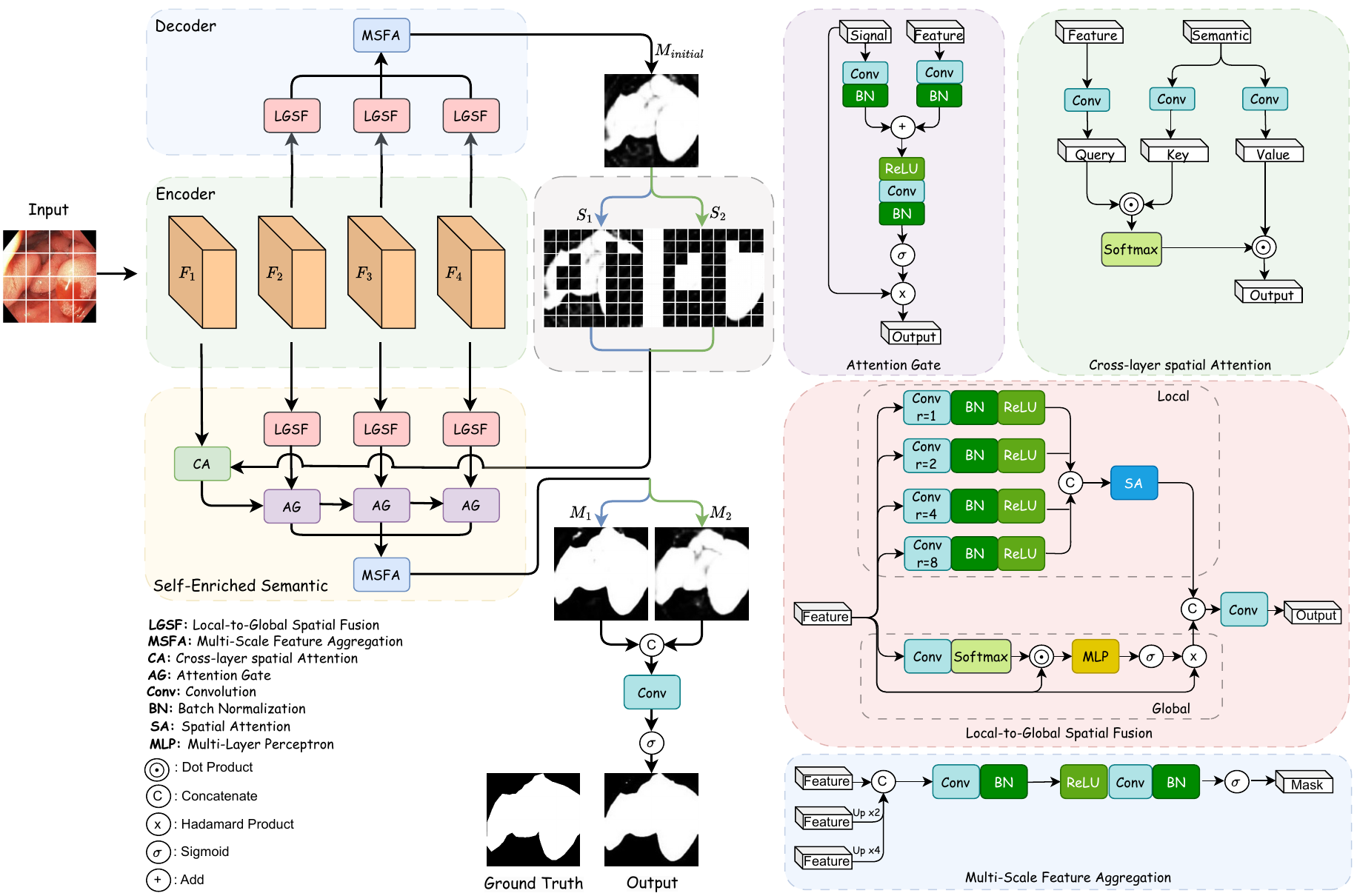}
    \end{center}
    \caption{Overview of our architecture. The proposed method consists of an Encoder (Section \ref{sec:encoder}), a Decoder (Section \ref{sec:decoder}) and a Self-Enriched Semantic (Section \ref{sec:semantic}) module. The Encoder extracts a sequence of multi-scale features from an input image. The Decoder aggregates high-level features to generate an initial segmentation mask. The Self-Enriched Semantic provides supplementary semantics to high-level features to relocate polyp objects.
}
    \label{figure2}
    \end{figure}
\end{center}

As depicted in Fig.~\ref{figure2}, our automatic polyp segmentation solution contains three principal components: Encoder, Decoder, and Self-Enriched Semantic (SES). The first is the Encoder, pretrained on ImageNet~\cite{imagenet}, extracts multi-scale features from an input image. The second contribution is the Decoder which employs Local-to-Global Spatial Fusion (LGSF) to capture both the global and local spatial features to achieve robust feature representation. Subsequently, refined features are aggregated to locate polyp objects and generate an initial global feature map. In the end, the SES component queries potential semantics from the initial global feature map and send them to high-level features to detect and relocate polyp objects accurately. 

\subsection{Encoder Backbone}
\label{sec:encoder}
In computer vision tasks, the encoder plays a crucial role in capturing spatial information and contextual cues from input images. Transformer-based encoding methods~\cite{Polyp-PVT, Transfuse, Transnetr} offer the ability to capture long-range dependence information across different areas within the input image. Metaformer~\cite{Metaformer} has recently introduced new insights into designing transformer architecture and has shown significant performance improvements in various computer vision tasks. Motivated by these findings, our study adopts a vision metaformer encoder known as Caformer~\cite{Metaformer} as a reliable and competitive backbone for feature extraction. Given an input image $I \in \mathbb{R}^{H \times W \times 3}$, the encoder extracts four levels of features denoted as $\{F_i| (i \in (1,2,3,4)\}$. Among these feature maps, $F_1$  provides detailed appearance information, while $F_2$, $F_3$, and $F_4$ offer high-level features.
\begin{equation}
F_{i}=\varphi_{\mathrm{Caformer}}(I) \in \mathbb{R}^{\frac{H}{2^{i+1}}\times \frac{W}{2^{i+1}}\times C_i}
\end{equation}

Where $H$, $W$, $C$ represent the height, width spatial, channel dimensions, respectively. In practice, we set $H$ and $W$ to 352, $C_i \in (64, 128, 320, 512)$.
\subsection{Global Feature Map Aggregation}
\label{sec:decoder}
The encoder features represent crucial and distinctive information essential for detecting polyp objects. Local features capture intricate details and boundaries, whereas global features provide contextual insights and spatial relationships among various structures. To effectively capture both global and local spatial information, we propose Local-to-Global Spatial Fusion (LGSF), as illustrated in Fig.~\ref{figure2}.

The local stage conducts four parallel dilated convolutions~\cite{dilated} with a dilation rate of $\{1, 2, 4, 8\}$ to extract local features at various spatial scales. Each dilated convolution is followed by batch normalization (BN) and a rectified linear unit (ReLU). Resultant features from four dilated convolutions are aggregated to obtain the local feature representation. The resulting feature representation is then processed by a spatial attention mechanism (SA)~\cite{sa} to suppress the irrelevant regions. The detail of the process is provided below:
\begin{equation}
    \textbf{F}_{local} = SA(Concatenate(C^{r=1}_{3\times3}(\textbf{F}), C^{r=2}_{3\times3}(\textbf{F}), C^{r=4}_{3\times3}(\textbf{F}), C^{r=8}_{3\times3}(\textbf{F})))
\end{equation}

The global stage incorporates non-local operation~\cite{non-local} to explore the long-range relationships between each pixel in the spatial space. This stage applies a convolution layer to obtain a feature representation. The resulting feature representation is transposed and followed by a Softmax function, and a Hadamard operation on the input feature $F$ to create a pixel relationship context. This context is then passed through an MLP layer to enhance the relationship representation. In the end, the resulting representation is followed by a sigmoid ($\sigma$) function and another Hadamard operation on the input feature $F$ to attain the global feature representation. The global stage can be delineated as follows:
\begin{equation}
\textbf{F}_{global} = \textbf{F} \times \sigma(MLP(\textbf{F} \odot Softmax((C_{1\times1}(\textbf{F}))^T))) 
\end{equation}

The final feature representations are obtained by combining local and global information, followed by a convolution layer. Multi-Scale Feature Aggregation (MSFA) module is used to synthesize multi-scale feature representations. This module fuses refined high-level features through a process as depicted in Fig.~\ref{figure2}.The first two features, $F_3$ and $ F_4$ undergo bilinear upsampling to match the spatial dimensions of all three features before they are concatenated. To enhance the capture of non-linear features, we further employ a series of convolutional layers, BN and ReLU. Finally, a sigmoid ($\sigma$) function is conducted to produce the output $M_{initial}$, which serves as the initial global feature map.

\subsection{Self-Enriched Semantic}
\label{sec:semantic}
The shallow layer features are closer to the input more than deep layer features, they preserve more of the original image's details and structure. Therefore, we leverage the low-level features $F_1$ to query implicit semantics from the initial global feature map, thereby providing supplementary semantics to the deep features. The semantic-enriched deep features are then decoded to yield two semantic-enriched segmentation, $M_1$ and $M_2$. The detailed structure of the Self-Enriched Semantic (SES) module is displayed in Fig.~\ref{figure2}. The process can be formulated as following: 

\begin{equation}
{\bf M}_{i} = MSFA({\bf F}^{rich}_2, {\bf F}^{rich}_3, {\bf F}^{rich}_4)
\end{equation}
\begin{equation}
{\bf M} = \sigma(C_{1\times1}(Concatenate({\bf M}_{1}, {\bf M}_{2})))
\end{equation}

Firstly, we consider the distribution of pixel values in patch-level images to represent the initial global feature map $M_{initial}$ to two distinct types of semantic areas including $S_1$ and $S_2$. Considering patch-level images where exist polyp objects, $S_1$ includes patches with pixel values varying from $0$ and $1$, often indicating noise or ambiguous boundaries that have not been sufficiently explored, whereas $S_2$ consists of patches with pixel values closer to $1$, representing solid structures of polyp objects. This categorization helps us differentiate between areas of interest and those that may introduce variability or noise. We then employ the $F_1$ as query and the $S_1$ acts as key-value pairs, applying Cross-layer spatial Attention (CA)\cite{transformer} to ascertain the relevance of $F_1$ and $S_1$. The resultant feature is then sent to high-level features through Attention Gate (AG) units progressively. In the end, we fuse semantic-enriched high-level features using Multi-Scale Fusion Aggregation (MSFA) to achieve a semantic-enriched global feature map, $M_{1}$. By Implementing the same operation to $F_1$ and $S_2$, we also obtain $M_{2}$. Finally, we concatenate $M_{1}$ and $M_{2}$ before passing them through a convolution layer followed by a sigmoid ($\sigma$) function to predict the final global feature map, $M$. 
\section{Experiments}
\subsection{Dataset and Evaluation Metrics}
Following recent cutting-edge solutions for the polyp segmentation task, we employ five widely-used benchmark datasets to assess the efficacy of our proposed model. These datasets include Kvasir~\cite{Kvasir}, ClinicDB~\cite{Clinic}, ColonDB~\cite{T1}, ETIS~\cite{Etis}, and EndoScene~\cite{Endoscene}. Table~\ref{tab1} provides a comprehensive overview of each dataset, including their specific usage details and objectives.

We employ various standard metrics to assess and compare the performance of polyp segmentation algorithms. The Dice score quantifies the spatial agreement between the predicted segmentation mask and the ground truth mask, whereas the IoU score computes the ratio of their overlapping area to the combined area. Both scores range between $0$ and $1$, with higher values indicating better segmentation performance. The MAE computes the average absolute difference between individual pixels in the predicted and ground truth masks. These evaluation metrics offer a comprehensive assessment of the segmentation performance, considering both spatial alignment and pixel-level precision.

\subsection{Implementation Details}
We utilize the power of RTX $3090$ GPU to accelerate both the training and inference stages of our model. Throughout the training process, we monitor various metrics including loss function, mDice, mIoU, and MAE scores to assess the performance and guide the training process. The total duration of training amounts to approximately $2$ hours to achieve optimal performance. Detailed training parameters are provided in Table~\ref{tab2}.
\begin{table}[t]
\caption{Specific usage details and objectives of Kvasir, ClinicDB, ColonDB, ETIS and EndoScene Datasets.}
\begin{center}
\begin{tabular}{lccccc}
\hline
Dataset & Image& Size& Train & Test & Objective \\
\hline
Kvasir & 1000 & Variable &  900  & 100 & Learning\\

ClinicDB & 612 & 384 × 288 & 550  & 62 & Learning\\

ColonDB & 380 & 574 × 500 & - & 380 & Generalization \\

ETIS & 196 & 1225 × 966 & - & 196 & Generalization\\

EndoScene & 60 & 574 × 500 & - & 60 & Generalization\\
\hline
\end{tabular}
\label{tab1}
\end{center}
\end{table}
\begin{table}[t]
\scriptsize	
\caption{Parameters of training configuration.}
\begin{center}
\begin{tabular}{lccc}
\hline
Image Size &Batch Size& Epoch & Loss Function \\

$352 \times 352$ & 16 & 200  & wBCE~\cite{bceloss} + wDice~\cite{diceloss}\\
\hline
Optimizer & Learning Rate & Decay Rate& Weight Decay \\

AdamW~\cite{adamw} & 1e-4 & 1e-1 &  1e-4  \\
\hline

\end{tabular}
\label{tab2}
\end{center}
\end{table}
\subsection{Comparisons with State-of-the-art Methods}
This section conducts a comprehensive evaluation focusing on two critical aspects: Learning ability, which verifies the segmentation performance on the seen dataset, and generalization ability, which evaluates the capacity of the model to generalize effectively to unseen data. A total of sixteen state-of-the-art models from the domain of the polyp segmentation, including U-Net~\cite{U-Net}, UNet++~\cite{Unet++}, PraNet~\cite{PraNet}, SFA~\cite{SFA}, MSEG~\cite{MSEG}, ACSNet~\cite{ACSNet}, DCRNet~\cite{DCRNet}, EU-Net~\cite{EU-Net} and SANet~\cite{SANet}, alongside newer models such as Polyp-PVT~\cite{Polyp-PVT}, ADSNet~\cite{ADSNet}, CaraNet~\cite{Cara}, TransUnet~\cite{TransUnet}, Transfuse~\cite{Transfuse}, UCTransNet~\cite{UctransNet}, SSFormer~\cite{SSFormer}, are collected for comparative analysis. The performance of these models is meticulously evaluated on five benchmark datasets using mDice, mIoU, and Mean Absolute Error (MAE) scores. In order to ensure fairness and reproducibility in our comparative analysis, we meticulously maintained consistency across training, validation, and testing datasets for all assessed models. Following the methodology outlined in PraNet~\cite{PraNet}, we adopt an identical dataset configuration as illustrated in Table~\ref{tab1}, comprising $900$ and $548$ images sourced from the Kvasir and ClinicDB datasets as the training set, with the remaining $64$ and $100$ images allocated as the respective test set to evaluate the learning ability. Additionally, we utilize the ColonDB, ETIS, and EndoScene datasets, which were not included in the training phase, to assess generalization ability.
\begin{table}[t]
\scriptsize	
\caption{Learning ability of diverse polyp segmentation baselines on Kvasir \& ClinicDB datasets across mDice, mIoU, MAE scores. ↑ denotes higher the better and ↓ denotes lower the better. \textbf{Bold} denotes the best score among the models, and \underline{underline} denotes the second best.}
\begin{center}
\begin{tabular}{lcccccc}
\toprule

Methods & \multicolumn{3}{c}{Kvasir}&\multicolumn{3}{c}{ClinicDB} \\

& \multicolumn{1}{c}{mDice↑}& \multicolumn{1}{c}{mIoU↑}& \multicolumn{1}{c}{MAE↓} & \multicolumn{1}{c}{mDice↑}& \multicolumn{1}{c}{mIoU↑}& \multicolumn{1}{c}{MAE↓} \\

\midrule
U-Net~\cite{U-Net} & \multicolumn{1}{c}{0.818} & \multicolumn{1}{c}{0.746} &  \multicolumn{1}{c}{0.055}  & \multicolumn{1}{c}{0.823 } & \multicolumn{1}{c}{0.755}&  \multicolumn{1}{c}{ 0.019}\\

UNet++~\cite{Unet++} & \multicolumn{1}{c}{0.821 } & \multicolumn{1}{c}{0.743} &  \multicolumn{1}{c}{ 0.048}  & \multicolumn{1}{c}{0.794 } & \multicolumn{1}{c}{0.729}&  \multicolumn{1}{c}{0.022}\\

SFA~\cite{SFA} & \multicolumn{1}{c}{0.723} & \multicolumn{1}{c}{ 0.611} &  \multicolumn{1}{c}{0.075}  & \multicolumn{1}{c}{0.700 } & \multicolumn{1}{c}{0.607}&  \multicolumn{1}{c}{ 0.042}\\

MSEG~\cite{MSEG} & \multicolumn{1}{c}{0.897 } & \multicolumn{1}{c}{ 0.839} &  \multicolumn{1}{c}{ 0.028}  & \multicolumn{1}{c}{0.909 } & \multicolumn{1}{c}{0.864}&  \multicolumn{1}{c}{\underline{0.007}}\\

DCRNet~\cite{DCRNet} & \multicolumn{1}{c}{0.886 } & \multicolumn{1}{c}{0.825} &  \multicolumn{1}{c}{0.035}  & \multicolumn{1}{c}{0.896 } & \multicolumn{1}{c}{0.844}&  \multicolumn{1}{c}{ 0.010}\\

ACSNet~\cite{ACSNet}  &\multicolumn{1}{c}{0.898} & \multicolumn{1}{c}{ 0.838} &  \multicolumn{1}{c}{ 0.032}  & \multicolumn{1}{c}{0.882 } & \multicolumn{1}{c}{0.826}&  \multicolumn{1}{c}{0.011}\\

PraNet~\cite{PraNet} & \multicolumn{1}{c}{0.898 } & \multicolumn{1}{c}{0.840} &  \multicolumn{1}{c}{0.030 }  & \multicolumn{1}{c}{0.899} & \multicolumn{1}{c}{ 0.849}&  \multicolumn{1}{c}{0.009}\\

EU-Net~\cite{EU-Net} & \multicolumn{1}{c}{0.908 } & \multicolumn{1}{c}{0.854} &  \multicolumn{1}{c}{0.028}  & \multicolumn{1}{c}{0.902} & \multicolumn{1}{c}{ 0.846}&  \multicolumn{1}{c}{ 0.011}\\

SANet~\cite{SANet} & \multicolumn{1}{c}{ 0.904 } & \multicolumn{1}{c}{0.847} &  \multicolumn{1}{c}{0.028}  & \multicolumn{1}{c}{0.916} & \multicolumn{1}{c}{ 0.859}&  \multicolumn{1}{c}{0.012}\\

Polyp-PVT~\cite{Polyp-PVT} & \multicolumn{1}{c}{ 0.917 } & \multicolumn{1}{c}{0.864} &  \multicolumn{1}{c}{\underline{0.023}}  & \multicolumn{1}{c}{0.937 } & \multicolumn{1}{c}{ 0.889}&  \multicolumn{1}{c}{\textbf{0.006}}\\

ADSNet~\cite{ADSNet} & \multicolumn{1}{c}{\underline{0.920}} & \multicolumn{1}{c}{\underline{0.871}} &  \multicolumn{1}{c}{\textbf{0.020}}  & \multicolumn{1}{c}{0.938} & \multicolumn{1}{c}{0.890}&  \multicolumn{1}{c}{\textbf{0.006}}\\

CaraNet~\cite{Cara} & \multicolumn{1}{c}{0.918 } & \multicolumn{1}{c}{0.865} &  \multicolumn{1}{c}{\underline{0.023}}  & \multicolumn{1}{c}{0.936  } & \multicolumn{1}{c}{0.887}&  \multicolumn{1}{c}{\underline{0.007}}\\

TransUnet~\cite{TransUnet} &\multicolumn{1}{c}{0.913  } & \multicolumn{1}{c}{0.857} &  \multicolumn{1}{c}{0.028}  & \multicolumn{1}{c}{0.935  } & \multicolumn{1}{c}{0.887}&  \multicolumn{1}{c}{0.008}\\

TransFuse~\cite{Transfuse} & \multicolumn{1}{c}{\underline{0.920}} & \multicolumn{1}{c}{0.870} &  \multicolumn{1}{c}{\underline{0.023}}  & \multicolumn{1}{c}{\underline{0.942}} & \multicolumn{1}{c}{\underline{0.897}} &  \multicolumn{1}{c}{\underline{0.007}}\\

UCTransNet~\cite{UctransNet} & \multicolumn{1}{c}{0.918 } & \multicolumn{1}{c}{0.860} &  \multicolumn{1}{c}{\underline{0.023}}  & \multicolumn{1}{c}{0.933} & \multicolumn{1}{c}{ 0.860}&  \multicolumn{1}{c}{ 0.008}\\
\midrule
\textbf{Polyp-SES} &\multicolumn{1}{c}{\textbf{0.924}} & \multicolumn{1}{c}{\textbf{0.875}} & \multicolumn{1}{c}{\textbf{0.020}} & \multicolumn{1}{c}{\textbf{0.945}} & \multicolumn{1}{c}{\textbf{0.902}} & \multicolumn{1}{c}{\textbf{0.006}} \\
\bottomrule
\end{tabular}
\label{tab3}
\end{center}
\end{table}
\begin{table*}[htbp]
\scriptsize	
\caption{Generalization ability of diverse polyp segmentation baselines on ColonDB, ETIS \& EndoScene datasets across mDice, mIoU, MAE scores. ↑ denotes higher the better and ↓ denotes lower the better. \textbf{Bold} denotes the best score among the models, and \underline{underline} denotes the second best.}
\begin{center}
\begin{tabular}{lccccccccc}
\toprule
Methods&\multicolumn{3}{c}{ColonDB}&\multicolumn{3}{c}{ETIS}&\multicolumn{3}{c}{EndoScene} \\

\textbf{} & \multicolumn{1}{c}{mDice↑}& \multicolumn{1}{c}{mIoU↑}& \multicolumn{1}{c}{MAE↓} & \multicolumn{1}{c}{mDice↑}& \multicolumn{1}{c}{mIoU↑}& \multicolumn{1}{c}{MAE↓} & \multicolumn{1}{c}{mDice↑}& \multicolumn{1}{c}{mIoU↑}& \multicolumn{1}{c}{MAE↓}\\

\midrule
U-Net~\cite{U-Net} & \multicolumn{1}{c}{0.512 } & \multicolumn{1}{c}{0.444} & \multicolumn{1}{c}{0.061} &\multicolumn{1}{c}{0.398} &\multicolumn{1}{c}{0.335} &\multicolumn{1}{c}{0.036}&\multicolumn{1}{c}{0.710} &\multicolumn{1}{c}{0.627}&\multicolumn{1}{c}{0.022}  \\

UNet++~\cite{Unet++} & \multicolumn{1}{c}{0.483} & \multicolumn{1}{c}{ 0.410} & \multicolumn{1}{c}{0.064  } &\multicolumn{1}{c}{0.401} &\multicolumn{1}{c}{0.344} &\multicolumn{1}{c}{0.035}&\multicolumn{1}{c}{0.707 } &\multicolumn{1}{c}{0.624}&\multicolumn{1}{c}{0.018}  \\

SFA~\cite{SFA} & \multicolumn{1}{c}{0.469 } & \multicolumn{1}{c}{0.347} & \multicolumn{1}{c}{0.094} &\multicolumn{1}{c}{0.297} &\multicolumn{1}{c}{ 0.217} &\multicolumn{1}{c}{ 0.109}&\multicolumn{1}{c}{0.467 } &\multicolumn{1}{c}{0.329}&\multicolumn{1}{c}{0.065}  \\

MSEG~\cite{MSEG} & \multicolumn{1}{c}{0.735 } & \multicolumn{1}{c}{0.666} & \multicolumn{1}{c}{0.038  } &\multicolumn{1}{c}{0.700} &\multicolumn{1}{c}{0.630} &\multicolumn{1}{c}{0.015}&\multicolumn{1}{c}{0.874 } &\multicolumn{1}{c}{0.804}&\multicolumn{1}{c}{0.009}  \\

DCRNet~\cite{DCRNet} & \multicolumn{1}{c}{0.704 } & \multicolumn{1}{c}{0.631} & \multicolumn{1}{c}{ 0.052} &\multicolumn{1}{c}{0.556} &\multicolumn{1}{c}{ 0.496} &\multicolumn{1}{c}{ 0.096}&\multicolumn{1}{c}{0.856 } &\multicolumn{1}{c}{0.788}&\multicolumn{1}{c}{ 0.010}  \\

ACSNet~\cite{ACSNet}  & \multicolumn{1}{c}{0.716} & \multicolumn{1}{c}{ 0.649} & \multicolumn{1}{c}{0.039 } &\multicolumn{1}{c}{0.578} &\multicolumn{1}{c}{ 0.509} &\multicolumn{1}{c}{0.059}&\multicolumn{1}{c}{0.863 } &\multicolumn{1}{c}{0.787}&\multicolumn{1}{c}{0.013}  \\

PraNet~\cite{PraNet} & \multicolumn{1}{c}{0.712} & \multicolumn{1}{c}{ 0.640} & \multicolumn{1}{c}{0.043 } &\multicolumn{1}{c}{ 0.628} &\multicolumn{1}{c}{0.567} &\multicolumn{1}{c}{0.031}&\multicolumn{1}{c}{0.871 } &\multicolumn{1}{c}{0.797}&\multicolumn{1}{c}{0.010}  \\

EU-Net~\cite{EU-Net} & \multicolumn{1}{c}{0.756} & \multicolumn{1}{c}{ 0.681} & \multicolumn{1}{c}{0.045} &\multicolumn{1}{c}{0.687 } &\multicolumn{1}{c}{0.609} &\multicolumn{1}{c}{0.067}&\multicolumn{1}{c}{0.837} &\multicolumn{1}{c}{ 0.765}&\multicolumn{1}{c}{0.015}  \\

SANet~\cite{SANet} & \multicolumn{1}{c}{0.753 } & \multicolumn{1}{c}{0.670} & \multicolumn{1}{c}{0.043 } &\multicolumn{1}{c}{ 0.750} &\multicolumn{1}{c}{0.654} &\multicolumn{1}{c}{0.015}&\multicolumn{1}{c}{0.888 } &\multicolumn{1}{c}{0.815}&\multicolumn{1}{c}{ 0.008}  \\

Polyp-PVT~\cite{Polyp-PVT} & \multicolumn{1}{c}{0.808} & \multicolumn{1}{c}{ 0.727} & \multicolumn{1}{c}{0.031  } &\multicolumn{1}{c}{0.787} &\multicolumn{1}{c}{0.706} &\multicolumn{1}{c}{0.013}&\multicolumn{1}{c}{0.900 } &\multicolumn{1}{c}{0.833}&\multicolumn{1}{c}{\underline{0.007}}  \\

ADSNet~\cite{ADSNet} & \multicolumn{1}{c}{\underline{0.815}} & \multicolumn{1}{c}{\underline{0.730}} & \multicolumn{1}{c}{\underline{0.029}} &\multicolumn{1}{c}{\underline{0.798}} &\multicolumn{1}{c}{0.715} &\multicolumn{1}{c}{\underline{0.012}}&\multicolumn{1}{c}{0.890} &\multicolumn{1}{c}{0.819}&\multicolumn{1}{c}{0.010}  \\

CaraNet~\cite{Cara} & \multicolumn{1}{c}{0.773} & \multicolumn{1}{c}{0.689} & \multicolumn{1}{c}{0.042} &\multicolumn{1}{c}{0.747} &\multicolumn{1}{c}{0.672} &\multicolumn{1}{c}{0.017}&\multicolumn{1}{c}{\underline{0.903}} &\multicolumn{1}{c}{\underline{0.838}}&\multicolumn{1}{c}{\underline{0.007}}  \\

TransUnet~\cite{TransUnet} & \multicolumn{1}{c}{0.781} & \multicolumn{1}{c}{0.699 } & \multicolumn{1}{c}{0.036} &\multicolumn{1}{c}{0.731} &\multicolumn{1}{c}{\underline{0.824}} &\multicolumn{1}{c}{ 0.021}&\multicolumn{1}{c}{0.893} &\multicolumn{1}{c}{0.660}&\multicolumn{1}{c}{0.009}  \\

TransFuse~\cite{Transfuse} & \multicolumn{1}{c}{0.781} & \multicolumn{1}{c}{0.706} & \multicolumn{1}{c}{0.035 } &\multicolumn{1}{c}{0.737} &\multicolumn{1}{c}{\textbf{0.826}} &\multicolumn{1}{c}{0.020}&\multicolumn{1}{c}{0.894} &\multicolumn{1}{c}{0.654}&\multicolumn{1}{c}{0.009}  \\

SSFormer~\cite{SSFormer} & \multicolumn{1}{c}{0.772} & \multicolumn{1}{c}{0.697} & \multicolumn{1}{c}{0.036} &\multicolumn{1}{c}{ 0.767} &\multicolumn{1}{c}{0.698 } &\multicolumn{1}{c}{0.016 }&\multicolumn{1}{c}{0.887 } &\multicolumn{1}{c}{0.821}&\multicolumn{1}{c}{\underline{0.007}}  \\
\midrule
\textbf{Polyp-SES} & \multicolumn{1}{c}{\textbf{0.817}} & \multicolumn{1}{c}{\textbf{0.741}}  & \multicolumn{1}{c}{\textbf{\textbf{0.026}}} & \multicolumn{1}{c}{\textbf{0.805}} & \multicolumn{1}{c}{0.756} & \multicolumn{1}{c}{\textbf{0.011}} & \multicolumn{1}{c}{\textbf{0.911}} & \multicolumn{1}{c}{\textbf{0.847}} & \multicolumn{1}{c}{\textbf{0.005}} \\
\bottomrule
\end{tabular}
\label{tab4}
\end{center}
\end{table*}
\begin{center}
    \begin{figure*}[t]
    \begin{center}
     \includegraphics[width=0.8\linewidth]{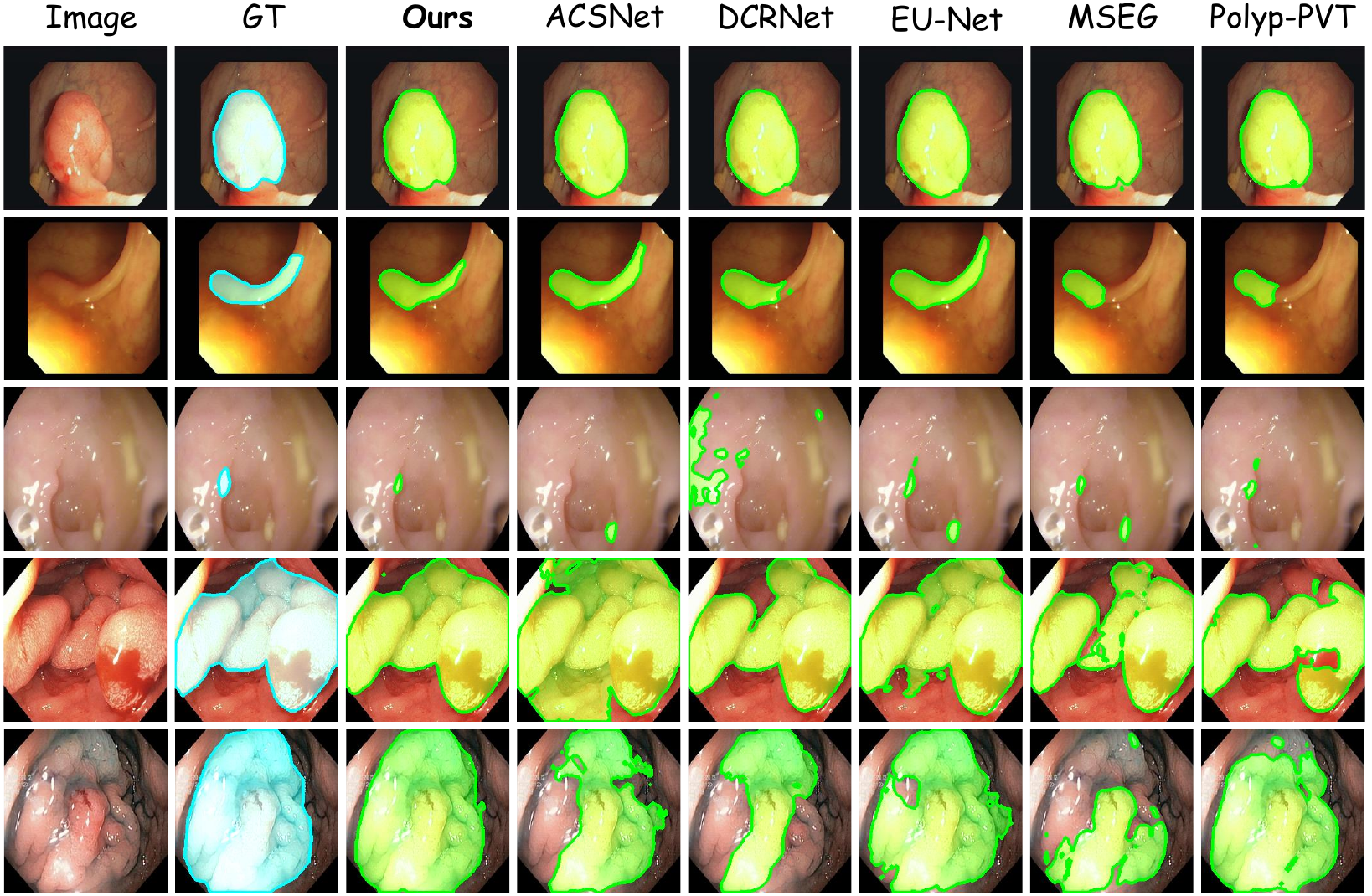}
    \end{center}
    \caption{Qualitative results with the current polyp segmentation baselines. Green indicates a predicted mask. It can be found, our proposed
model can precisely recognize and segment polyp objects even under the variability in polyp appearance attached to noises, ambiguous boundaries, and intricate foregrounds.}
    \label{figure5}
    \end{figure*}
\end{center}
\begin{center}
    \begin{figure*}[htbp]
    \begin{center}
     \includegraphics[width=0.8\linewidth]{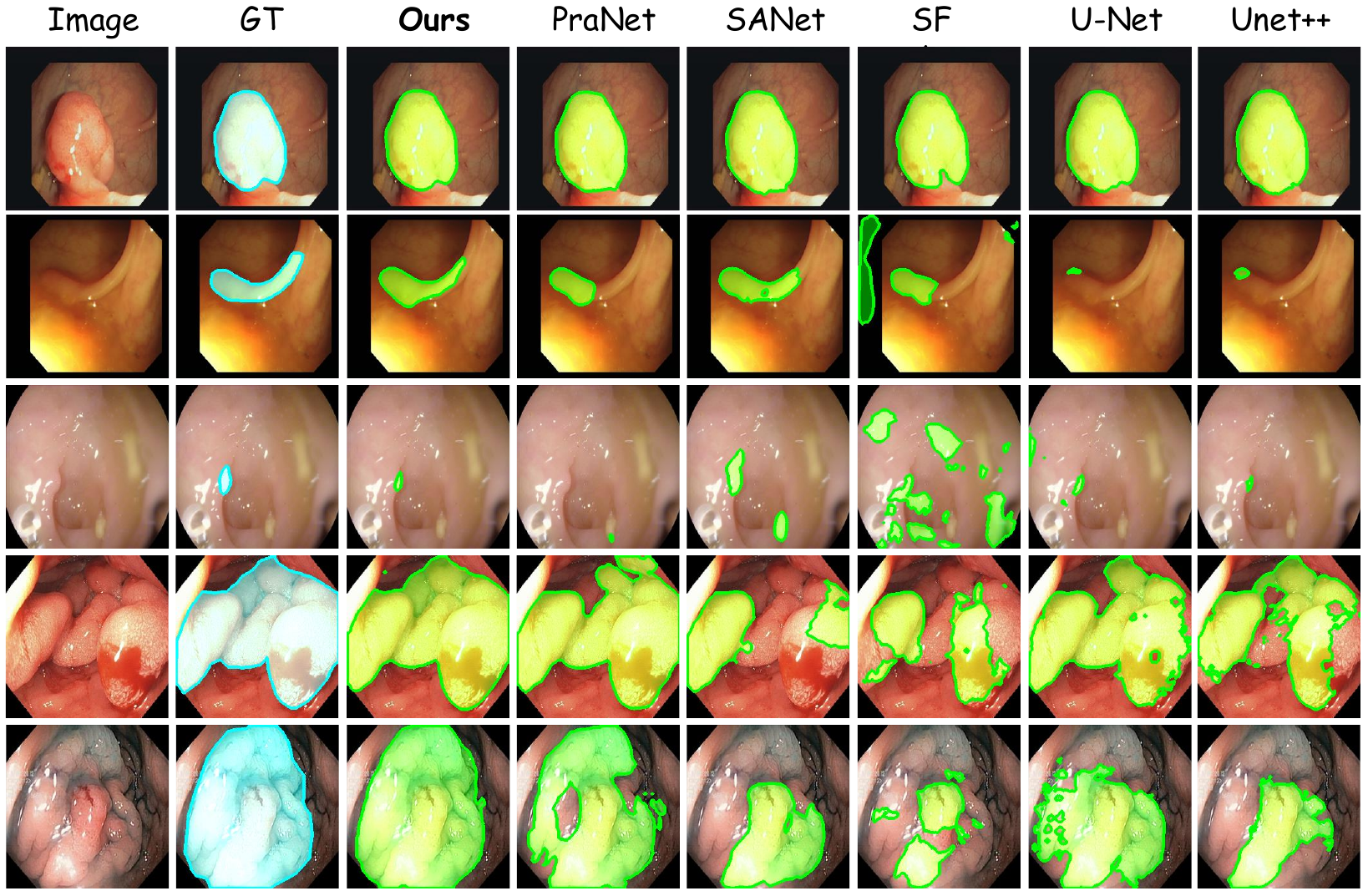}
    \end{center}
    \caption{Qualitative results with the current polyp segmentation baselines. Green indicates a predicted mask. It can be found, our proposed
model can precisely recognize and segment polyp objects even under the variability in polyp appearance attached to noises, ambiguous boundaries, and intricate foregrounds.}
    \label{figure6}
    \end{figure*}
\end{center}
\textbf{Learning ability.} In the learning ability experiment, the domain of the test and train set is similar. Table~\ref{tab3} presents the results of different cutting-edge models on the Kvasir and ClinicDB datasets. Our method demonstrates outstanding performance compared to recently published models on both datasets, as evidenced by the mDice, mIoU, and MAE scores. Specifically, our method obtains a mDice score of $0.924$, a mIoU score of $0.875$ on the Kvasir dataset, outperforming the second-best model ADSNet~\cite{ADSNet}. On the ClinicDB dataset, our model achieves a mDice score and mIoU of $0.945$ and $0.902$, respectively, showcasing an improvement compared to TransFuse~\cite{Transfuse}. These results underscore the robustness and effectiveness of the proposed method in terms of learning ability.

\textbf{Generalization ability.} We conduct a through evaluation of the polyp segmentation baselines to assess their generalization performance on unseen datasets, as shown in Table~\ref{tab4}. It can be observed that our method demonstrates competitive performance across all three datasets compared to other techniques. Specifically, our model is higher than the second-best ADSNet~\cite{ADSNet} on the ColonDB dataset in term of mDice score and mIoU score. On ETIS dataset, although Transfuse~\cite{Transfuse} exhibits notable performance with a mIoU score of $0.826$, its corresponding mDice score is lower at $0.737$. In contrast, our results achieve a mDice score of $0.805$, outperforming all other models, alongside a mIoU score of $0.756$. These findings highlight the stable performance of our proposed approach, which excels in both mDice and mIoU scores where other methods may have limitations. Additionally, our model demonstrates remarkable improvement on the EndoScene dataset, with mDice score, mIoU score, and MAE score of $0.911$, $0.847$, and $0.005$, respectively. These results underscore the superior generalization capability of our proposed method.

\textbf{Qualitative results.}
We present qualitative results comparing our model with other polyp segmentation baselines across five datasets, depicted in Fig.~\ref{figure5} and Fig.~\ref{figure6}. The segmentation results of the compared methods are sourced from the publicly available Polyp-PVT~\cite{Polyp-PVT}. We can observe that our model produces clear and precise segmentation outcomes across a variety of polyp structures. Furthermore, it effectively identifies and segments polyp objects under different variations in image quality, minimizing artifacts and extraneous regions while maintaining exceptional segmentation accuracy. These findings underscore the efficiency and accuracy of our proposed segmentation algorithm, even in challenging spatial scenarios where previous methods have struggled.
\subsection{Ablation Study}
In the ablation study section, we conduct experiments to validate the necessity and effectiveness of each proposed module in the overall architecture individually. Our standard polyp segmentation architecture includes an Encoder, Decoder and Self-Enriched Semantic (SES). The ablation studies are conducted on all five polyp datasets, evaluating based on mDice and mIoU scores.
\subsubsection{Effectiveness of Encoder Backbone}
In the first ablation study, we assess the effectiveness of different encoder backbones. We use the proposed standard architecture as the baseline and replace diverse encoder backbones, consisting of ResNet50~\cite{ResNet50} (CNN), PVT~\cite{PVT} (Transformer), and Caformer~\cite{Metaformer} (Metaformer). All variants are trained under the same configuration, and the results are summarized in Table~\ref{tab5}. It is evident that the standard baseline, with Caformer as the encoder backbone, achieves superior performance with higher mDice and mIoU scores across all five datasets compared to CNN-based or conventional transformer encoder backbones. This demonstrates the effectiveness of exploiting the vision metaformer as encoder backbone in extracting robust features and enhancing polyp segmentation performance.
\subsubsection{Effectiveness of Local-to-Global Spatial Fusion} 
\begin{center}
    \begin{figure*}[t]
    \begin{center}
     \includegraphics[width=0.75\linewidth]{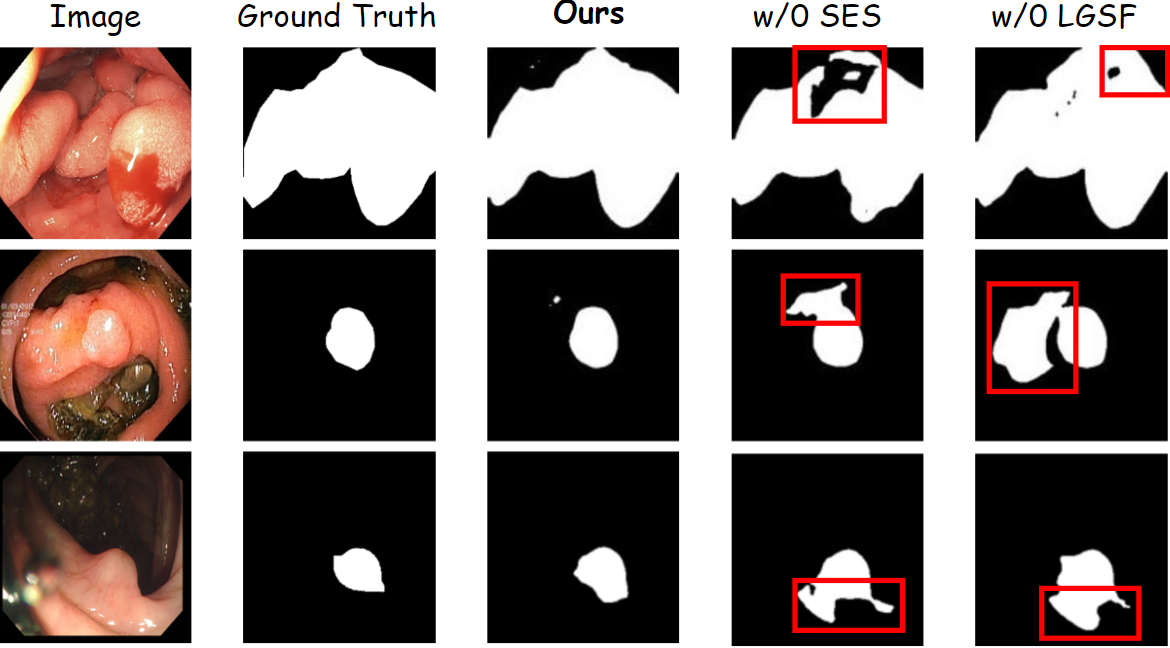}
    \end{center}
    \caption{Visualization of the ablation study results. As can be seen,
removing Self-Enrich Semantic (SES) leads to segmentation failures in challenging semantic areas, whereas the removal of Local-to-Global Spatial Fusion (LGSF) causes incorrectly segmentation results denoted as red-bordered boxes.}
    \label{figure7}
    \end{figure*}
\end{center}

\begin{center}
    \begin{figure*}[t]
    \begin{center}
     \includegraphics[width=.75\linewidth]{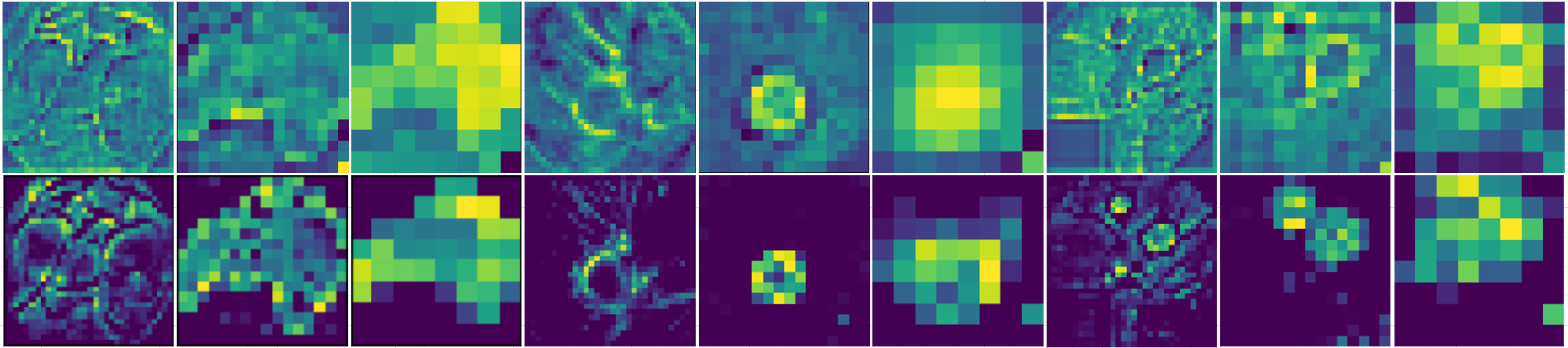}
    \end{center}
    \caption{Visualization of the high-level feature maps before and after refining by the Local-to-Global Spatial Fusion (LGSF). The first row is high-level features, the second row is features which are refined by LGSF. As can be observed, the LGSF captures helpful characteristics of level-specific features and removes redundant information from other regions.}
    \label{figure8}
    \end{figure*}
\end{center}

To assess the impact of local and global feature aggregation, we remove the LGSF units from the decoder in the standard architecture, and replace them with $3\times3$ convolution layers. Results presented in Table~\ref{tab6} demonstrate a significant decrease in both mDice and mIoU scores compared to the standard baseline with LGSF units. Furthermore, visualizations of segmentation predictions in Fig.~\ref{figure7} reveal that the absence of LGSF introduces considerable noise. These qualitative and quantitative results prove that LGSF can help model to distinguish polyp tissues and contribute greatly to the polyp segmentation performance. In order to further explore the contribution of the LGSF, we showcase high-level features before and after refinement by the LGSF units in Fig.~\ref{figure8}. As can be observed, the LGSF eliminate redundant information from other regions and yield informative characteristics of level-specific features, aiding the model in precisely locating polyp objects and enhancing segmentation performance.
\begin{table*}[t]
\scriptsize	
\caption{Ablation study of various encoder backbones over five benchmarks. ↑ denotes higher the better. \textbf{Bold} denotes the best score.}
\begin{center}
\resizebox{\textwidth}{!}{%
\begin{tabular}{lccccccccccc}
\toprule
Encoder& Type &\multicolumn{2}{c}{Kvasir}&\multicolumn{2}{c}{ClinicDB}&\multicolumn{2}{c}{ColonDB}&\multicolumn{2}{c}{ETIS}&\multicolumn{2}{c}{EndoScene} \\
\textbf{} & & \multicolumn{1}{c}{mDice↑}& mIoU↑& \multicolumn{1}{c}{mDice↑} & mIoU↑ & \multicolumn{1}{c}{mDice↑} & mIoU↑ & \multicolumn{1}{c}{mDice↑} & mIoU↑& \multicolumn{1}{c}{mDice↑} & mIoU↑ \\

\midrule
ResNet50~\cite{ResNet50} & CNN & \multicolumn{1}{c}{0.909} & \multicolumn{1}{c}{0.852} & \multicolumn{1}{c}{0.932} & \multicolumn{1}{c}{0.880} & \multicolumn{1}{c}{0.797} & \multicolumn{1}{c}{0.722} & \multicolumn{1}{c}{0.804} & \multicolumn{1}{c}{0.727} & \multicolumn{1}{c}{0.895} & \multicolumn{1}{c}{0.827}\\

PVT~\cite{PVT} & Transformer & \multicolumn{1}{c}{0.919} & \multicolumn{1}{c}{0.870} & \multicolumn{1}{c}{0.933} & \multicolumn{1}{c}{0.884} & \multicolumn{1}{c}{0.804} & \multicolumn{1}{c}{0.726} & \multicolumn{1}{c}{0.779} & \multicolumn{1}{c}{0.695} & \multicolumn{1}{c}{0.892} & \multicolumn{1}{c}{0.826}\\

Caformer~\cite{Metaformer} & Metaformer & \multicolumn{1}{c}{\textbf{0.924}} & \multicolumn{1}{c}{\textbf{0.875}} & \multicolumn{1}{c}{\textbf{0.945}} & \multicolumn{1}{c}{\textbf{0.902}} & \multicolumn{1}{c}{\textbf{0.817}} & \multicolumn{1}{c}{\textbf{0.741}} & \multicolumn{1}{c}{\textbf{0.805}} & \multicolumn{1}{c}{\textbf{0.756}} & \multicolumn{1}{c}{\textbf{0.911}} & \multicolumn{1}{c}{\textbf{0.847}}\\
\bottomrule
\end{tabular}}
\label{tab5}
\end{center}
\end{table*}
\begin{table*}[t]
\scriptsize	
\caption{Ablation study of LGSF and SES over five benchmarks. ↑ denotes higher the better. \textbf{Bold} denotes the best score}
\begin{center}
\begin{tabular}{lcccccccccc}
\toprule
Method&\multicolumn{2}{c}{Kvasir}&\multicolumn{2}{c}{ClinicDB}&\multicolumn{2}{c}{ColonDB}&\multicolumn{2}{c}{ETIS} &\multicolumn{2}{c}{EndoScene}\\

\textbf{} & \multicolumn{1}{c}{mDice↑}& mIoU↑& \multicolumn{1}{c}{mDice↑} & mIoU↑ & \multicolumn{1}{c}{mDice↑} & mIoU↑ & \multicolumn{1}{c}{mDice↑} & mIoU↑ & \multicolumn{1}{c}{mDice↑} & mIoU↑ \\

\midrule
w/o SES, LGSF & \multicolumn{1}{c}{0.900} & \multicolumn{1}{c}{0.850} & \multicolumn{1}{c}{0.909} & \multicolumn{1}{c}{0.862} & \multicolumn{1}{c}{0.775} & \multicolumn{1}{c}{0.699} & \multicolumn{1}{c}{0.691} & \multicolumn{1}{c}{0.615} & \multicolumn{1}{c}{0.891} & \multicolumn{1}{c}{0.819}\\

w/o SES & \multicolumn{1}{c}{0.905} & \multicolumn{1}{c}{0.853} & \multicolumn{1}{c}{0.923} & \multicolumn{1}{c}{0.874} & \multicolumn{1}{c}{0.784} & \multicolumn{1}{c}{0.708} & \multicolumn{1}{c}{0.729} & \multicolumn{1}{c}{0.654} & \multicolumn{1}{c}{0.888} & \multicolumn{1}{c}{0.811}\\

w/o LGSF & \multicolumn{1}{c}{ 0.918} & \multicolumn{1}{c}{0.869} & \multicolumn{1}{c}{0.912} & \multicolumn{1}{c}{0.868} & \multicolumn{1}{c}{0.781} & \multicolumn{1}{c}{0.694} & \multicolumn{1}{c}{0.786} & \multicolumn{1}{c}{0.702} & \multicolumn{1}{c}{0.888} & \multicolumn{1}{c}{0.824}\\
\midrule
\textbf{Ours} & \multicolumn{1}{c}{\textbf{0.924}} & \multicolumn{1}{c}{\textbf{0.875}} & \multicolumn{1}{c}{\textbf{0.945}} & \multicolumn{1}{c}{\textbf{0.902}} & \multicolumn{1}{c}{\textbf{0.817}} & \multicolumn{1}{c}{\textbf{0.741}} & \multicolumn{1}{c}{\textbf{0.805}} & \multicolumn{1}{c}{\textbf{0.756}} & \multicolumn{1}{c}{\textbf{0.911}} & \multicolumn{1}{c}{\textbf{0.847}}\\
\bottomrule
\end{tabular}
\label{tab6}
\end{center}
\end{table*}
\subsubsection{Effectiveness of Self-Enriched Semantic} 
\begin{center}
    \begin{figure*}[t]
    \begin{center}
     \includegraphics[width=0.75\linewidth]{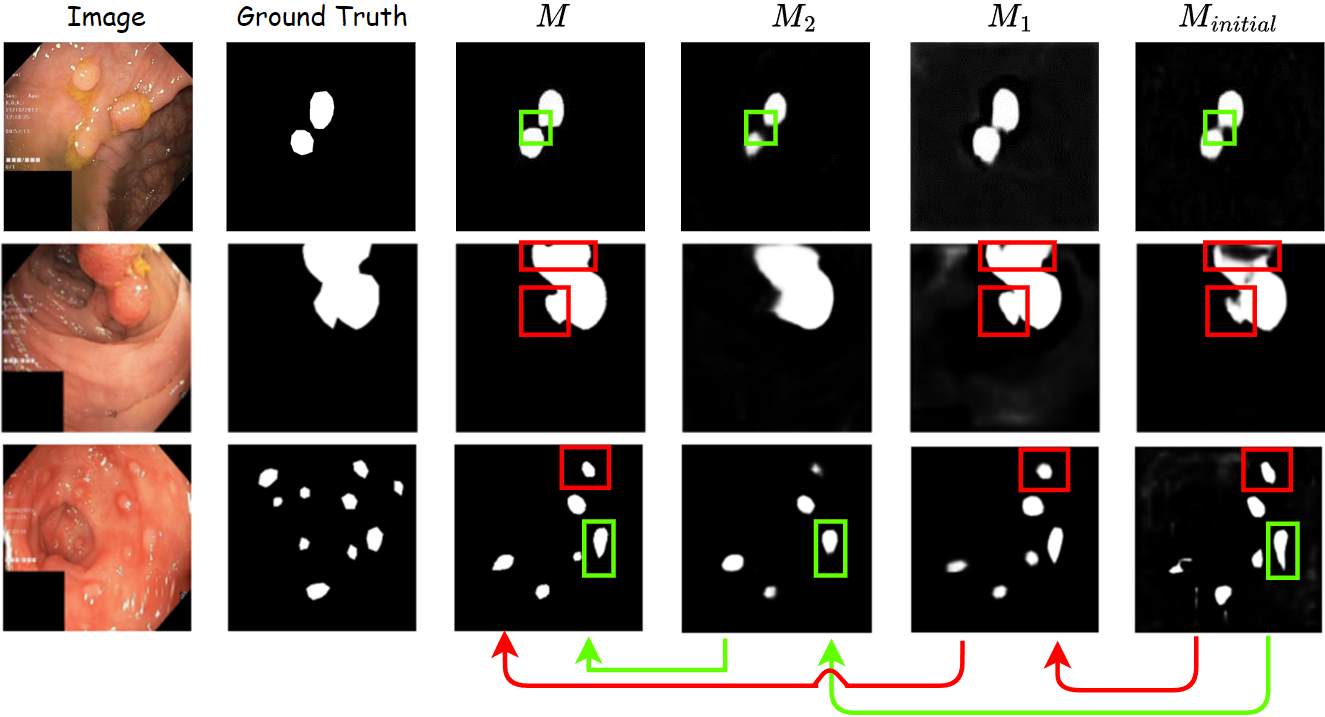}
    \end{center}
    \caption{Visualization of $M_{initial}$, $M_{1}$, $M_{2}$ and $M$ predictions. It is found that the final global feature map $M$ are constructed and contributed by two semantic-enriched segmentation masks $M_{1}$, and $M_{2}$. For example, in the \textbf{first} row, $M_{2}$ separates successfully two close polyp objects. In the \textbf{second} row, $M_{1}$ explores missing semantic areas to reconstruct the feature map. In the \textbf{third} row, $M_{1}$ and $M_{2}$ relocate tiny polyp objects more precisely.}
    \label{figure9}
   \end{figure*}
\end{center}

This ablation study validates the effectiveness of the proposed SES module on the overall architecture. By excluding the SES module from the baseline, we revert to a conventional encoder-decoder structure. The performance presented in Table~\ref{tab6} reveals that the conventional encoder-decoder architecture without SES leads to a deterioration in performance, with lower on mDice score and mIoU score compared our standard model. In Fig.~\ref{figure7}, it is apparent that the absence of the SES results in more detailed errors or missed semantic areas. This proves that the SES module facilitates the model to explore potential semantics to give the better global feature map with the comprehensive context. We further investigate the contribution of the SES by visualizing the two semantic-enriched segmentation masks containing $M_{1}$ and $M_{2}$ in Fig.~\ref{figure9}. Notably, $M_{1}$ demonstrates the ability to explore potential semantic areas referring to regions denoted as red-bordered boxes where were not previously captured by $M_{initial}$. Meanwhile, $M_{2}$ concerns the solid structural components of polyp objects in green-bordered boxes where already captured by $M_{initial}$. Taking advantage of $M_{1}$ and $M_{2}$, we attain a final global feature map with comprehensive semantics, thereby improving polyp segmentation performance.
\section{Conclusion}
In this paper, we introduce ``Automatic Polyp Segmentation with Self-Enriched Semantic Model'', an innovative approach aimed at addressing the limitations of contemporary methods in capturing comprehensive contexts. By leveraging a vision metaformer Encoder, a Decoder, and a Self-Enriched Semantic module, our method effectively enriches deep features with supplementary semantics, improving the model's understanding of challenging contexts. Through quantitative and qualitative experiments, we demonstrate its effectiveness and superiority over state-of-the-art models across five polyp benchmarks, evaluated on mDice, mIoU, and MAE metrics, showcasing its proficiency in both learning and generalization abilities. Additionally, we conducted through studies to understand the underlying reasons for its effectiveness, offering valuable insights that can guide future research in medical image segmentation-related tasks, particularly those focused on automatic polyp segmentation.
\section*{Acknowledgements}
This work was supported by Institute of Information \& communications Technology Planning \& Evaluation (IITP) under the Artificial Intelligence Convergence Innovation Human Resources Development (IITP-2023-RS-2023-00256629) grant funded by the Korea government(MSIT). This research was supported by the MSIT(Ministry of Science and ICT), Korea, under the ITRC(Information Technology Research Center) support program(IITP-2024-RS-2024-00437718) supervised by the IITP(Institute for Information \& Communications Technology Planning \& Evaluation). This study was supported by a grant (HCRI 23038) Chonnam National University Hwasun Hospital Institute  for Biomedical Science. The corresponding author is Soo-Hyung Kim.

%
%
\bibliographystyle{splncs04}
\bibliography{main}
\end{document}